\documentclass[11pt]{article}

\usepackage[preprint]{acl}

\usepackage{times}
\usepackage{latexsym}

\usepackage[T1]{fontenc}

\usepackage[utf8]{inputenc}

\usepackage{microtype}

\usepackage{inconsolata}

\usepackage{graphicx}

\usepackage{booktabs}
\usepackage{pdfpages}
\usepackage{enumitem}
\usepackage{multirow}
\usepackage{makecell}
\usepackage{color}
\usepackage{bm}
\usepackage{bbding}
\usepackage{subcaption}
\usepackage{colortbl}
\usepackage{tcolorbox}
\usepackage{amsmath}
\usepackage{pifont}
\usepackage{wrapfig}

\usepackage{marvosym}
\usepackage{graphicx}
\usepackage[normalem]{ulem}
\usepackage{CJKutf8}
\usepackage{tocloft}
\usepackage{amssymb}
\usepackage{algorithm}
\usepackage{algpseudocode}

\newcommand{\ie}{\textit{i.e.}}

\newcommand{\modelname}{DPWriter}

\title{DPWriter: Reinforcement Learning with \\Diverse Planning Branching for Creative Writing}

\author{
  Qian Cao\textsuperscript{\textmd{1}}\footnotemark[2]\thanks{Work done during an internship at Kuaishou Technology.\\
    \indent\ \ \textsuperscript{$\dagger$} Equal contribution.\\
    \indent\ \ \textsuperscript{\Letter} Corresponding author.
  },
  Yahui Liu\textsuperscript{\textmd{2}}\footnotemark[2],
  Wei Bi\textsuperscript{\textmd{2}},
  Yi Zhao\textsuperscript{\textmd{1}},
  Ruihua Song\textsuperscript{\textmd{1}\Letter},
  Xiting Wang\textsuperscript{\textmd{1}\Letter}, \\
  \textbf{Ruiming Tang\textsuperscript{\textmd{2}},
  Guorui Zhou\textsuperscript{\textmd{2}},
  Han Li\textsuperscript{\textmd{2}}}
  \\
  $^1$Renmin University of China, $^2$Kuaishou Technology \\
  \{caoqian4real, rsong, xitingwang\}@ruc.edu.cn, \\
  yahui.cvrs@gmail.com\\
}

\begin{document}

\maketitle

\begin{abstract}

Reinforcement learning (RL)-based enhancement of large language models (LLMs) often leads to reduced output diversity, undermining their utility in open-ended tasks like creative writing.
Current methods lack explicit mechanisms for guiding diverse exploration and instead prioritize optimization efficiency and performance over diversity.
This paper proposes an RL framework structured around a semi-structured long Chain-of-Thought (CoT), in which the generation process is decomposed into explicitly planned intermediate steps.
We introduce a Diverse Planning Branching method that strategically introduces divergence at the planning phase based on diversity variation, alongside a group-aware diversity reward to encourage distinct trajectories. 
Experimental results on creative writing benchmarks demonstrate that our approach significantly improves output diversity without compromising generation quality, consistently outperforming existing baselines.

\end{abstract}

\section{Introduction}

Diversity is a fundamental characteristic of the world and a core manifestation of human creativity~\cite{fischer2005distances,Bruce2008Difference}.
As large language models (LLMs) continue to advance in reasoning capabilities~\cite{chen2025towards}, and more recently by reinforcement learning (RL)~\cite{schulman2017proximal,shao2024deepseekmath,lu2025writing,Adithya2025RLMT}, the diversity exhibited in their generated texts has become particularly important.
However, users may experience a significant loss of content diversity when collaborating with LLMs for creative writing~\cite{Vishakh2024Does}.
This issue is even more pronounced in models trained with reinforcement learning from human feedback (RLHF)~\cite{o2024attributing}.

To alleviate the diversity decline caused by RL training~\cite{Robert2024UnderstandingRLHF,Vishakh2024Does,shypula2025evaluating}, a growing body of works are proposed~\cite{he2025rewarding,Li2025darling,anschel2025gapo}, yet several challenges remain in effectively enhancing diversity for LLMs.
First, many approaches focus on modifying reward functions~\cite{Jens2025RepresentationBased,he2025rewarding,Li2025darling}.
However, these methods largely leave the rollout process unconstrained, providing limited control over how diverse trajectories are explored during RL.
Second, some methods investigate branching or forking strategies to explore diverse trajectories~\cite{zheng2025fr3e,li2025treepo,guo2025spo}, but they primarily focus on improving sample efficiency~\cite{zheng2025parallel,wen2025parathinker} or overall performance~\cite{liu2025ettrl}, rather than explicitly targeting diversity as an intrinsic objective.
Moreover, they typically branch rollouts from high-entropy tokens~\cite{wang2025beyond}, which makes the branching process less controllable.

\begin{figure}[t!]
    \centering
    \includegraphics[width=1.0\linewidth]{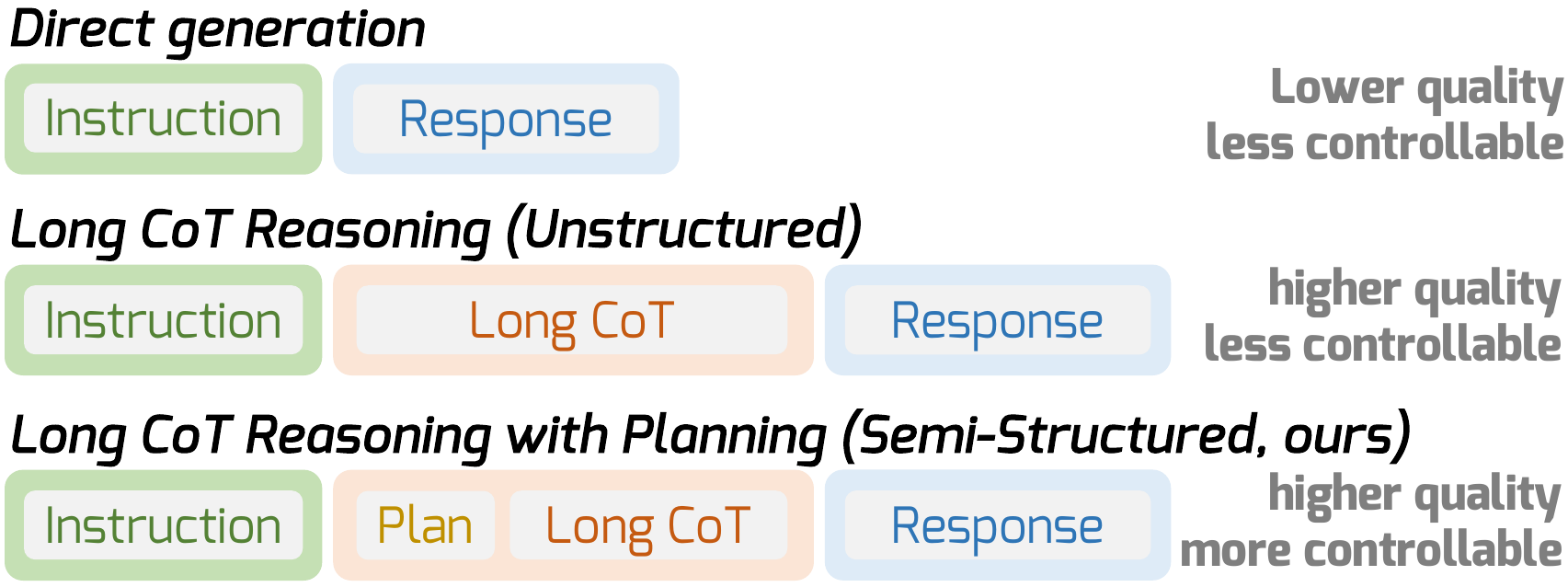}
    \caption{Comparison among three generation paradigms. Our semi-structured reasoning paradigm introduces global planning before reasoning, providing high-level guidance while maintaining higher quality.}
    \label{fig:paradigm_comparison}
\end{figure}

In this paper, we propose \modelname, which uses a semi-structured long Chain-of-Thought (CoT) as a scaffold to guide the RL process for improved diversity in LLMs.
As shown in Figure~\ref{fig:paradigm_comparison}, by decomposing the generation process into multiple stages, beginning with a global planning phase followed by long CoT reasoning and final response generation, our method provides explicit intermediate stages that facilitate diverse exploration.
Specifically, we introduce a Diverse Planning Branching (DPB) method that strategically branches diverse plans at our rollout-time DPB stage based on their diversity, allowing for more controlled and effective exploration of diverse trajectories.
We also incorporate a diversity reward that evaluates the diversity contribution of each response based on its group, collaborating with our rollout-time DPB strategy to further encourage diverse generation.
Extensive experiments on various creative writing benchmarks demonstrate that our approach significantly enhances the diversity of LLM-generated texts while maintaining high quality, consistently outperforming existing baselines.
The main contributions of this work are as follows:

$\bullet$ \textit{Diversity-guided RL framework.} We propose a novel RL framework that leverages semi-structured long CoT to guide the generation process. To support this, we construct a curated dataset comprising 43K writing instructions with semi-structured long CoT and high-quality responses.

$\bullet$ \textit{Planning-level diversity mechanisms.} We introduce a Diverse Planning Branching (DPB) method that strategically branches candidate plans at the planning stage, along with a diversity reward that evaluates each response according to its contribution within a group. This design enables controlled exploration and effectively promotes diverse generation trajectories.

$\bullet$ \textit{Experimental validation.} Experiments on multiple creative writing benchmarks demonstrate that our method significantly improves the diversity of LLM-generated texts while maintaining high generation quality, consistently outperforming existing baselines. Further analysis reveals that the DPB method and diversity reward work synergistically, jointly promoting more diverse generation.

\section{Related Work}
\label{sec:related_work}

\noindent\textbf{Diversity in Non-RL Training.}\quad
Previous studies have shown that supervised fine-tuning (SFT) or preference optimization may reduce output diversity~\cite{o2024attributing,Robert2024UnderstandingRLHF}, motivating another line of work aimed to alleviating this issue during training.
\citet{Li2025PreservingDiversity} emphasizes the overfitting issue inherent in SFT and introduces a game-theoretic framework to address the limitations of cross-entropy loss.
For preference optimization, some studies~\cite{Jack2025divpo,deshpande2025diverse} propose modifications to DPO~\cite{Rafael2023DPO} that focus on improved selection of diversified data samples.
Other methods~\cite{John2025ddpo,Abhijnan2025DPL} promote both output diversity and quality by employing weighted training objectives that better capture nuanced preferences.

\noindent\textbf{RL-based Methods for Diversity.}\quad
More recently, reinforcement learning (RL) has demonstrated strong effectiveness in improving model capabilities~\cite{wei2025igniting,Adithya2025RLMT}, leading to increased attention to methods that boost diversity during RL training.
A primary strategy modifies the reward in policy gradient methods like GRPO~\cite{shao2024deepseekmath} by introducing diversity-aware bonus~\cite{anschel2025gapo,Jens2025RepresentationBased} or penalty terms~\cite{Chen2025DRAGRPO,he2025rewarding,Li2025darling} , which are computed using diversity metrics over a group of generated responses.
The common goal is to shape the policy gradient to favor diverse and high-quality outputs.
In addition, some researchers have explored alternative approaches to adjust RL objectives, such as incorporating token-level entropy regularization~\cite{Yao2025DiversityAware}, designing semantic diversity terms~\cite{chen2025post}, or decoupling an entropy component from the KL divergence term~\cite{Stewart2025SPL}.

Although some studies investigate branching or forking strategies to enhance RL exploration~\cite{zheng2025fr3e,li2025treepo,guo2025spo}, they primarily aim to improve sample efficiency~\cite{zheng2025parallel,wen2025parathinker} and overall performance~\cite{liu2025ettrl} rather than explicitly promoting diversity.
Moreover, these works mainly focus on high-entropy tokens~\cite{wang2025beyond} as branching points~\cite{zheng2025fr3e,liu2025ettrl} or set fixed segment lengths for branching~\cite{li2025treepo,guo2025spo}, which makes the resulting rollouts less controllable.
Unlike prior work, our method integrates a semi-structured long CoT reasoning process with diversity-aware branching strategies to explicitly encourage the exploration of multiple, divergent planning pathways, making it suitable for open-ended creative writing tasks.

\section{Preliminaries}

\noindent\textbf{Task Formulation.}\quad
Given an instruction $q \sim \mathcal{Q}$ from an open-ended task like creative writing, the goal of a model $\mathcal{M}$ is to generate a response $y$ to the instruction, \ie, $y \sim \mathcal{M}(\cdot|q)$.
Using long CoT reasoning to generate a response can be formulated as first generating a reasoning chain $c$, followed by generating the final response, \ie, $c \sim \mathcal{M}(\cdot|q)$ and $y \sim \mathcal{M}(\cdot|q,c)$.
However, existing CoT reasoning processes are unstructured and implicitly learned, lacking explicit planning representations.

To allow high-level objectives to directly shape subsequent reasoning and final responses, we propose a semi-structured long CoT reasoning paradigm that introduces an explicit planning stage before reasoning.
Specifically, the model first generates a global plan $p$ for the response, then produces a reasoning chain $c$ conditioned on both the instruction and the plan, and finally generates the response conditioned on all of them, \ie, $p \sim \mathcal{M}(\cdot|q)$, $c \sim \mathcal{M}(\cdot|q,p)$ and $y \sim \mathcal{M}(\cdot|q,p,c)$.
In our proposed paradigm, as shown in Figure~\ref{fig:paradigm_comparison}, the plan $p$ serves as a high-level structural guide for the subsequent free-form reasoning and response generation.
This strikes a balance between the flexibility of unstructured reasoning chains and the control provided by explicit planning.

\noindent\textbf{RL for LLMs.}\quad
In the context of RL for LLMs, the model $\mathcal{M}$ is treated as a policy $\pi_\theta$ with parameters $\theta$, and the generation process is formulated as a Markov Decision Process (MDP).
At time step $t$, generating a token $a_t$ is treated as taking an action in state $s_t$, where the state consists of the instruction and all previously generated tokens \ie, $s_t = (q, a_1, a_2, \ldots, a_{t-1})$. 
In our semi-structured reasoning paradigm, the generation of a rollout $o$ can be denoted as the following distribution:
\begin{align}
    \pi_\theta(o \mid q) = 
    \prod\limits_{\tau \in \{p,c,y\}} 
    \prod\limits_{l=1}^{L_\tau} 
    \pi_\theta(a_l^{(\tau)} \mid q, \tau_{<l}),
    \label{loss_func}
\end{align}
where $\tau$ iterates over the plan $p$, reasoning chain $c$, and response $y$. $L_\tau$ is the length of sequence $\tau$, and $a_l^{(\tau)}$ is the $l$-th token in it.
The objective of RL is to maximize the expected cumulative reward:
\begin{align}
    J(\theta) = \mathbb{E}_{o \sim \pi_\theta(\cdot|q)} \left[ r(q,o) \right],
    \label{eq:rl_objective}
\end{align}
where $r(q,o)$ is the reward function that evaluates the quality of rollout $o$ given the instruction $q$.

\noindent\textbf{Group Relative Policy Optimization (GRPO).}\quad
\label{sec:grpo}
Our method is built upon GRPO~\citep{shao2024deepseekmath}, a recent RL algorithm that discards the critic model and estimates advantages within a group of $n$ rollouts $\{o_i\}_{i=1}^n$ generated by the old policy $\pi_{\theta_\text{old}}$ as in~\cite{schulman2017proximal}.
GRPO optimizes the policy $\pi_\theta$ by maximizing the following objective:
\begin{align}
    &J_\text{GRPO}(\theta) = \mathbb{E}_{q \sim \mathcal{Q}, \{o_i\}_{i=1}^N \sim \pi_{\theta_\text{old}}(\cdot|q)} \\
    &{\textstyle\frac{1}{n} \sum\limits_{i=1}^n \frac{1}{|o_i|} \sum\limits_{t=1}^{|o_i|} 
    \Bigl[
        \text{CLIP}(\rho_{i,t}, A_t) - \beta \mathbb{D}_{\text{KL}}(\pi_\theta || \pi_{\theta_\text{ref}})
    \Bigr]}, \nonumber 
    \label{eq:grpo_objective}
\end{align}
where $\rho_{i,t} = \frac{\pi_\theta(o_{i,t}|q,o_{i,<t})}{\pi_{\theta_\text{old}}(o_{i,t}|q,o_{i,<t})}$ is the importance sampling ratio at step $j$ of rollout $o_i$.
The clip function $\text{CLIP}(\rho_{i,t}, A_t) = \min(\rho_{i,t} A_t, \text{clip}(\rho_{i,t}, 1-\epsilon, 1+\epsilon) A_t)$ is used to limit the policy update step size~\cite{schulman2017proximal}, and the $\mathbb{D}_{\text{KL}}$ term penalizes the divergence from a reference policy $\pi_{\theta_\text{ref}}$ to further ensure stability, with $\beta$ being the penalty coefficient.
Given rewards $\{r_i\}_{i=1}^n$ of a group of rollouts $\{o_i\}_{i=1}^n$, the advantage $A_t$ for each rollout is computed as $A_t = \frac{r_i - \bar{r}}{\sigma_r}$, where $\bar{r}$ and $\sigma_r$ are the mean and standard deviation of the rewards.

\begin{figure}[t!]
    \centering
    \includegraphics[width=1.0\linewidth]{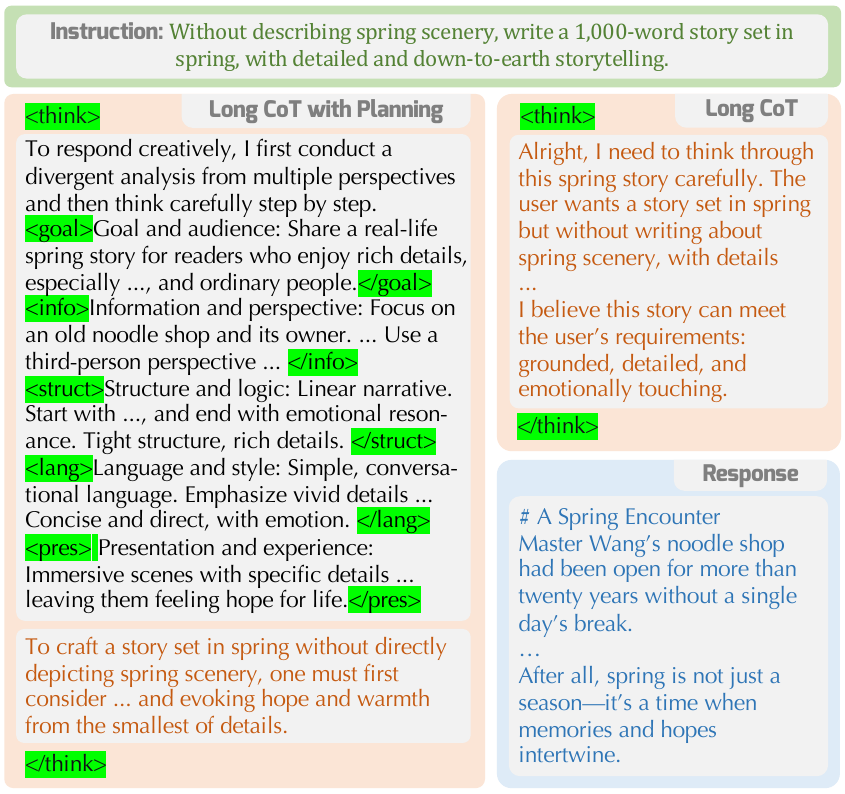}
    \caption{An example of the original long CoT data and the semi-structured long CoT with planning. Texts with \colorbox[HTML]{05FE04}{colored} background represent special tokens.}
    \label{fig:data_example}
\end{figure}

\begin{figure*}[t!]
    \centering
    \includegraphics[width=1.0\linewidth]{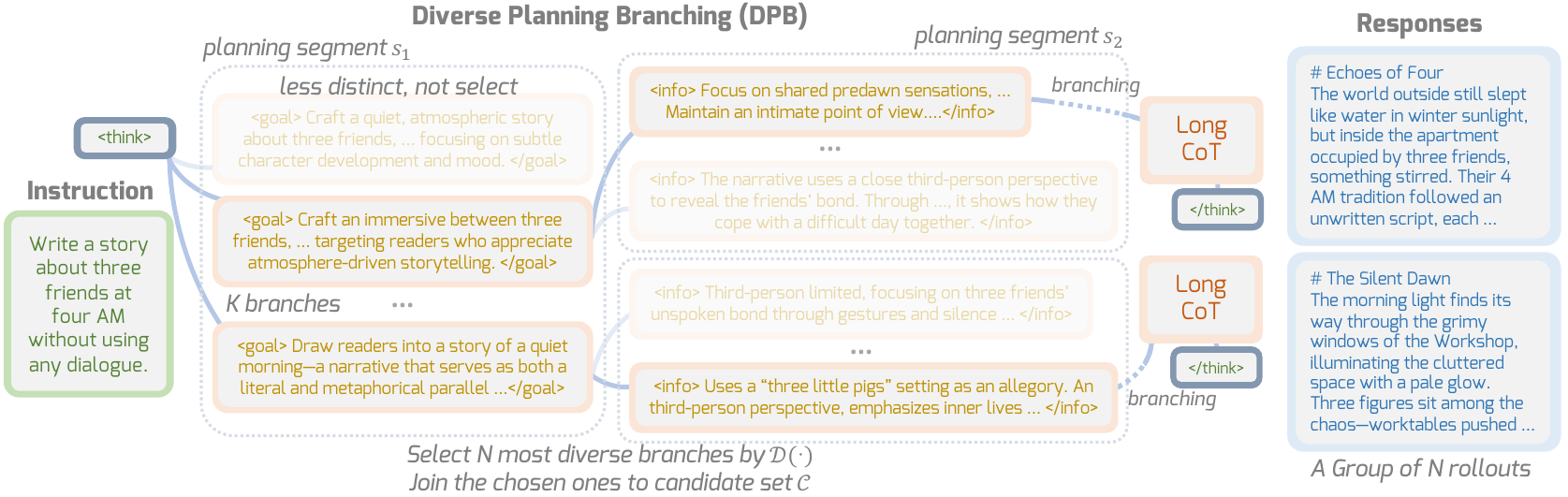}
    \caption{An overview of our Diverse Planning Branching method. During RL, at each planning segment, we branch out multiple diverse continuations from each candidate, forming a pool of candidates. We then select the most diverse ones to proceed to the next segment, ultimately generating diverse final responses.}
    \label{fig:dpb_overview}
\end{figure*}

\section{Constructing Semi-structured CoT Data with Planning}
\label{sec:data_const}

We present a two-step method for constructing semi-structured CoT data through multi-aspect planning and plan-consistent reasoning, enhancing coherence and controllability.

\noindent\textbf{Multi-aspect Planning Generation.}\quad
Given an input instruction $q$, a long CoT reasoning $c$, and a target response $y$, the goal is to generate a plan $p$ that outlines key aspects to guide both the CoT and response generation.
Inspired by rhetorical and writing theories~\cite{bitzer1968rhetorical,flower1981cognitive,spangher2025creative}, we design a multi-aspect planning framework that includes the following aspects:

\noindent
$\bullet$ \textit{Goal and Audience}: To define and identify the primary objective and target audience. \\
$\bullet$ \textit{Information and Perspective}: To highlight the key information to be included and the perspective or viewpoint to be adopted. \\
$\bullet$ \textit{Structure and Logic}: To outline the logical flow and structure of the response, including main points and their organization. \\
$\bullet$ \textit{Language and Style}: To specify the desired tone, vocabulary, and stylistic elements to be used. \\
$\bullet$ \textit{Presentation and Experience}: To describe how the information should be presented to enhance reader engagement and experience.

We employ GPT-4.1~\cite{openai2025gpt4-1} to produce the plan $p$ based on the instruction $q$ and response $y$.
Details of the used prompt are in Appendix~\ref{appendix:prompts}.

\noindent\textbf{Plan-consistent CoT Generation.}\quad
However, directly inserting the generated plan $p$ at the beginning of the long CoT $c$ may introduce inconsistency between the plan and the reasoning process.
To overcome this, we use the constructed plan $p$ to revise the original CoT $c$ into a plan-consistent CoT $c'$.
This is achieved by using GPT-4.1 to revise the original CoT $c$ based on the plan $p$, ensuring alignment with the planned aspects while preserving the original information to avoid information drift.
The revision prompt is provided in Appendix~\ref{appendix:prompts}.

We introduce special tokens such as $\langle\mathtt{goal}\rangle$, $\langle\mathtt{/goal}\rangle$, $\langle\mathtt{info}\rangle$ and $\langle\mathtt{/info}\rangle$, etc., to enclose individual aspects of the plan $p$, making the structure explicit and easier for the model to recognize and follow.
An example of the original data sample $(q, c, y)$ and the semi-structured CoT data sample $(q, p, c', y)$ is shown in Figure~\ref{fig:data_example}.

\section{Method}

In this section, we present \textit{Diverse Planning Branching} method and \textit{Rewarding Diversity Contribution} strategy, which exploit planning controllability in semi-structured CoTs to enhance rollout diversity and response quality.

\subsection{Diverse Planning Branching}

\noindent\textbf{Planning Capability Cold Start.}\quad
The initial semi-structured long CoT $(p, c')$ provides explicit planning cues and a consistent CoT reasoning process, both of which are reflected in the final response $y$.
To equip the model with this capability, we first cold-start the base model through supervised fine-tuning (SFT) on the semi-structured CoT data, allowing it to learn the planning formats and generate coherent CoT reasoning.
As shown in Section~\ref{sec:sft_ablation}, SFT using our semi-structured data yields performance comparable to or exceeding that obtained with other CoT datasets.

\noindent\textbf{Branching Planning Segments.}\quad
In the RL stage, the cold-started model acts as the policy model $\pi_\theta$.
Given an instruction $q$, the model first generates the planning part before producing the CoT $c'$ and response $y$.
By explicitly encouraging exploration over diverse planning strategies at each planning point, the policy induces diverse reasoning paths and final outputs. This, in turn, offers the reward model a broader set of high-quality candidates, improving both generation quality and diversity.

As shown in Figure~\ref{fig:dpb_overview}, we identify $S$ planning segments in the semi-structured CoT format, each segment $s$ is delimited by start and end tokens $(\mathtt{t}_s^\text{start}, \mathtt{t}_s^\text{end})$ (\textit{e.g.}, $\langle\mathtt{goal}\rangle$ and $\langle\mathtt{/goal}\rangle$).
During generation, for each planning segment $s$, we expand every candidate in the current candidate set $\mathcal{C}$ by sampling $K$ continuations starting from $\mathtt{t}_s^\text{start}$ until reaching the corresponding ending token $\mathtt{t}_s^\text{end}$.
Here, $K$ denotes the \textit{branch factor}, which controls the number of diverse continuations generated for each candidate.
This process yields a candidate pool of size $|\mathcal{C}| \times K$.
To select $G$ candidates for the next segment, where $G$ is the \textit{group size}, we measure candidate diversity using a predefined diversity metric $D(\cdot)$ and select the most diverse candidates.
For the first segment, the $G$ candidates are selected directly from the entire pool.
For subsequent segments, we select one candidate from each group of continuations originating from the same previous candidate, thereby ensuring diversity across different branches.
After processing all $S$ segments, we decode each candidate in $\mathcal{C}$ completion, yielding $B \times G$ final responses for a batch of $B$ instructions.

\noindent\textbf{Diversity Metrics.}\quad
To measure candidate diversity during branching, we consider two types of metrics:
(1) \textit{N-gram-based Diversity}, which calculates distinct n-grams across candidates to encourage lexical variety, 
and (2) \textit{semantic Diversity}, which measures the average pairwise cosine distance between candidate embeddings by using an \textit{off-the-shelf} embedding model (\ie, Qwen3-Embedding-0.6B~\cite{qwen3embedding}) to capture semantic differences.
These metrics jointly encourage exploration of diverse reasoning paths.

\begin{table*}[t!]
\centering
\scalebox{0.922}{
\setlength{\tabcolsep}{1.7mm}{
\begin{tabular}{p{5.8cm}ccccccccc}
\toprule \midrule
\multicolumn{1}{c}{\multirow{2}{*}{\textbf{Methods}}} &
  \multicolumn{3}{c}{\textbf{WritingBench}} &
  \multicolumn{3}{c}{\textbf{Creative Writing v3}} &
  \multicolumn{3}{c}{\textbf{ArenaHard v2.0}} \\ \cmidrule(lr){2-4} \cmidrule(lr){5-7} \cmidrule(lr){8-10}
\multicolumn{1}{c}{} &
  \multicolumn{1}{c}{Score} &
  \multicolumn{1}{c}{Emb} &
  \multicolumn{1}{c}{EAD} &
  \multicolumn{1}{c}{ELO} &
  \multicolumn{1}{c}{Emb} &
  \multicolumn{1}{c}{EAD} &
  \multicolumn{1}{c}{WR} &
  \multicolumn{1}{c}{Emb} &
  \multicolumn{1}{c}{EAD} \\ \midrule \midrule
Qwen3-4B-Base         & 3.74 & 15.84 & 5.94 & 43.88 & 33.21 & 5.20 & 1.9 & 33.27 & 6.26  \\ \midrule
Qwen3-4B~\cite{Yang2025Qwen3}   & 6.37 & 7.55 & 6.32 & 457.84 & 13.61 & 10.15 & 9.0 & 17.33 & 12.09  \\
GRPO~\cite{shao2024deepseekmath} & 6.32 & 9.07 & 8.02 & 659.83 & 17.00 & 15.67 & 11.0 & 22.27 & 15.82  \\
GRPO-Unlikeliness~\cite{he2025rewarding} & 6.28 & 9.46 & 8.33 & 660.46 & 17.07 & 15.15 & 12.1 & 22.79 & 15.92  \\
Darling~\cite{Li2025darling}     & 6.23 & 8.82 & 7.66 & 666.10 & 16.73 & 16.43 & 10.1 & 21.67 & 16.21  \\
GAPO~\cite{anschel2025gapo}      & 6.25 & 9.83 & 8.11 & 619.89 & 17.61 & 15.73 & 11.8 & 23.16 & 16.27  \\
\modelname\ (ours)               & \textbf{6.43} & \textbf{10.45} & \textbf{8.81} & \textbf{694.69} & \textbf{17.69} & \textbf{17.02} & \textbf{13.9} & \textbf{23.65} & \textbf{17.68}  \\

\midrule \midrule
Llama-3.2-3B-Instruct & 3.54 & 11.27 & 6.97 & 445.05 & 17.22 & 10.16 & 5.4 & 19.91 & 6.32  \\
\midrule
GRPO~\cite{shao2024deepseekmath} & 5.25 & 12.01 & 9.42 & 754.08 & 16.72 & 12.25 & 21.7 & 23.32 & 15.18  \\
GRPO-Unlikeliness~\cite{he2025rewarding} & 4.47 & 11.34 & 8.42 & 718.31 & 17.35 & 10.99 & 2.5 & \textbf{24.87} & 6.17  \\
Darling~\cite{Li2025darling}     & 4.57 & 9.31 & 7.79 & 759.05 & 14.21 & 12.40 & 19.5 & 21.25 & 15.18  \\
GAPO~\cite{anschel2025gapo}      & 4.57 & 10.65 & 8.05 & 730.24 & 15.97 & 12.19 & 20.5 & 23.57 & 15.37  \\
\modelname\ (ours)               & \textbf{5.31} & \textbf{12.03} & \textbf{9.60} & \textbf{829.05} & \textbf{17.72} & \textbf{12.50} & \textbf{29.0} & 22.56 & \textbf{15.45} \\

\midrule \bottomrule
\end{tabular}}}
\caption{Performance comparison of different methods on WritingBench, Creative Writing v3, and ArenaHard v2.0 (creative writing subset) benchmarks. The best results are \textbf{bolded}. `Emb' and `EAD' denote the embedding-based and ngram-based diversity metrics, respectively. `WR' denotes the win rate against gemini-2.0-flash.}
\label{tab:main_results}
\end{table*}

\subsection{Rewarding Quality and Diversity}
To encourage reasoning paths that yield both high-quality and diverse responses, we follow previous works~\cite{Chen2025DRAGRPO,he2025rewarding,Li2025darling} by jointly incorporating quality and diversity rewards during RL training.

\noindent\textbf{Quality Reward.}\quad
For the quality reward, we utilize a reward model $R_{\phi}$ trained on human preference data to assess the quality of generated responses.
Given a response $y_i$ to an instruction $q$, the quality reward is defined as:
\begin{equation}
    r^{\text{qua}}_{i}(q, y_i) = R_{\phi}(q, y_i)
\end{equation}
This reward encourages the model to generate responses that align with human preferences.

\noindent\textbf{Diversity Contribution Reward.}\quad
To further promote diversity among the generated responses, we introduce a \textit{Diversity Contribution Reward}, which measures how much a response contributes to the overall diversity of the response group.
The core intuition is to reward responses that introduce unique elements not shared by others, thereby promoting varied content generation.
Formally, given a response group $\mathcal{Y} = \{y_1, y_2, \dots, y_n\}$ for an instruction $q$, the diversity contribution reward for a response $y_i$ is defined as:
\begin{equation}
    r^{\text{div}}_{i}(q, y_i, \mathcal{Y}) = \text{Norm}\big(\frac{D(y_i, \mathcal{Y} \setminus \{y_i\})}{|y_i|}\big)
\end{equation}
where $D(\cdot)$ counts the unique n-grams in $y_i$ that do not appear in the other responses in $\mathcal{Y} \setminus \{y_i\}$, and \( |y_i| \) denotes the number of tokens in \( y_i \). The normalization function  \( \text{Norm}(\cdot) \) ensures the reward is on a comparable scale across responses.

To balance quality and diversity, we combine the two rewards as follows:
\begin{equation}
    r_{i}(q, y_i, \mathcal{Y}) = (1 - \lambda) \cdot r^{\text{qua}}_{i} + \lambda \cdot r^{\text{qua}}_{i} \cdot r^{\text{div}}_{i}
\end{equation}
where $\lambda\in[0, 1]$ controls the contribution of the diversity reward.
The diversity contribution reward is activated only when the response quality exceeds a certain threshold $\tau$ when $r^{\text{qua}}_{i} > \tau$; otherwise, we set $\lambda=0$.
This formulation ensures that responses are rewarded for diversity only when they satisfy a minimum quality threshold, favoring high-quality responses that also contribute meaningful diversity and guiding the model toward generating responses that are both high quality and diverse.

\section{Experiments}
In this section, we evaluate our proposed method on several benchmark datasets and compare it with relevant baselines.
We further conduct ablation studies to analyze key components of our approach and provide discussions of the results.

\subsection{Experimental Setup}
\noindent\textbf{Training Datasets.}\quad
We adopt open datasets of creative writing for training the model, including DeepWriting~\cite{wang2025reverse}, WritingPrompts~\cite{Fan18WP}, CreateSet~\cite{cao2025evaluating}, and COIG-Writer~\cite{li2025coig}.
Due to the large size of CreateSet and WritingPrompts, we randomly sample 13K and 12K examples from them, respectively.
For the data only containing instructions and responses, we generate long CoTs using GPT-4.1~\cite{openai2025gpt4-1} for them, where the prompts are provided in Appendix~\ref{appendix:prompts}.
After deduplication, the final dataset used to train the cold-start SFT model contains 43K samples.

For effective RL training, we apply data filtering to keep the samples on which the SFT model underperforms.
Specifically, we use the SFT model to generate responses for all training samples and score them with the reward model Skywork-Reward-V2-Llama-3.1-8B~\cite{liu2025skywork}, which is ranked first on RewardBench\footnote{https://huggingface.co/spaces/allenai/reward-bench}.
Samples whose maximum reward scores are lower than the overall average are retrained, resulting in 10K samples for RL training.

\noindent\textbf{Backbones and Baselines.}\quad
Our experiments are conducted on two different backbones, Qwen3-4B-Base~\cite{Yang2025Qwen3} and Llama-3.2-3B-Instruct~\cite{dubey2024llama}.
We compare our method with several strong baselines, including:

\noindent(1) \textit{GRPO}~\cite{shao2024deepseekmath}:
The standard GRPO described in Section~\ref{sec:grpo}; 

\noindent(2) \textit{GRPO-Unlikeliness}~\cite{he2025rewarding}:
A revised version of GRPO that rewards responses inversely to their likelihood where lower generation probability yields higher weight.

\noindent(3) \textit{Darling}~\cite{Li2025darling}:
A baseline combines a learned diversity classifier to calculate diversity reward from partitions.

\noindent(4) \textit{GAPO}~\cite{anschel2025gapo}:
An extension of GRPO enables models to learn distributional properties like uniform sampling.

\noindent
Our code is based on VeRL~\cite{Sheng2025verl} framework.
During RL training, the batch size is set to 128 with an update batch size of 32.
The group size $n$ is set to 8.
All the baselines are trained with the same data and settings as our method for fair comparison.

\noindent\textbf{Evaluation Benchmarks.}\quad
For evaluating both the quality and diversity of generated responses, we conduct experiments on three benchmarks, WritingBench~\cite{Wu2025WritingBench}, Creative Writing v3 (EQ-Bench)~\cite{Paech2023EQBench}, and ArenaHard v2.0 (creative writing subset)~\cite{Li2025ArenaHard}.
We report the average quality score (Score) of WritingBench, the normalized ELO score (stands for quality by win rate), and the win rate (WR) with style control on ArenaHard v2.0.
We further assess diversity using NoveltyBench~\cite{Zhang2025NoveltyBench} and report the Distinct metric, which partitions model outputs into equivalence classes based on a binary classifier trained to predict functional equivalence between generated samples.
We use Claude Sonnet 4~\cite{anthropic2025claude4} as the judge model for WritingBench and Creative Writing v3, and DeepSeek-V3~\cite{deepseek2024deepseekv3} for ArenaHard v2.0.
Diversity is measured by generating 16 responses per prompt and computing both the embedding-based average cosine distance (Emb) and the n-gram-based distinct. \ie, Expectation-Adjusted Distinct (EAD)~\cite{Liu2022ead} scores.
More details are in Appendix~\ref{appendix:implementation_details}.

\begin{figure}[t!]
    \centering
    \includegraphics[width=1.0\linewidth]{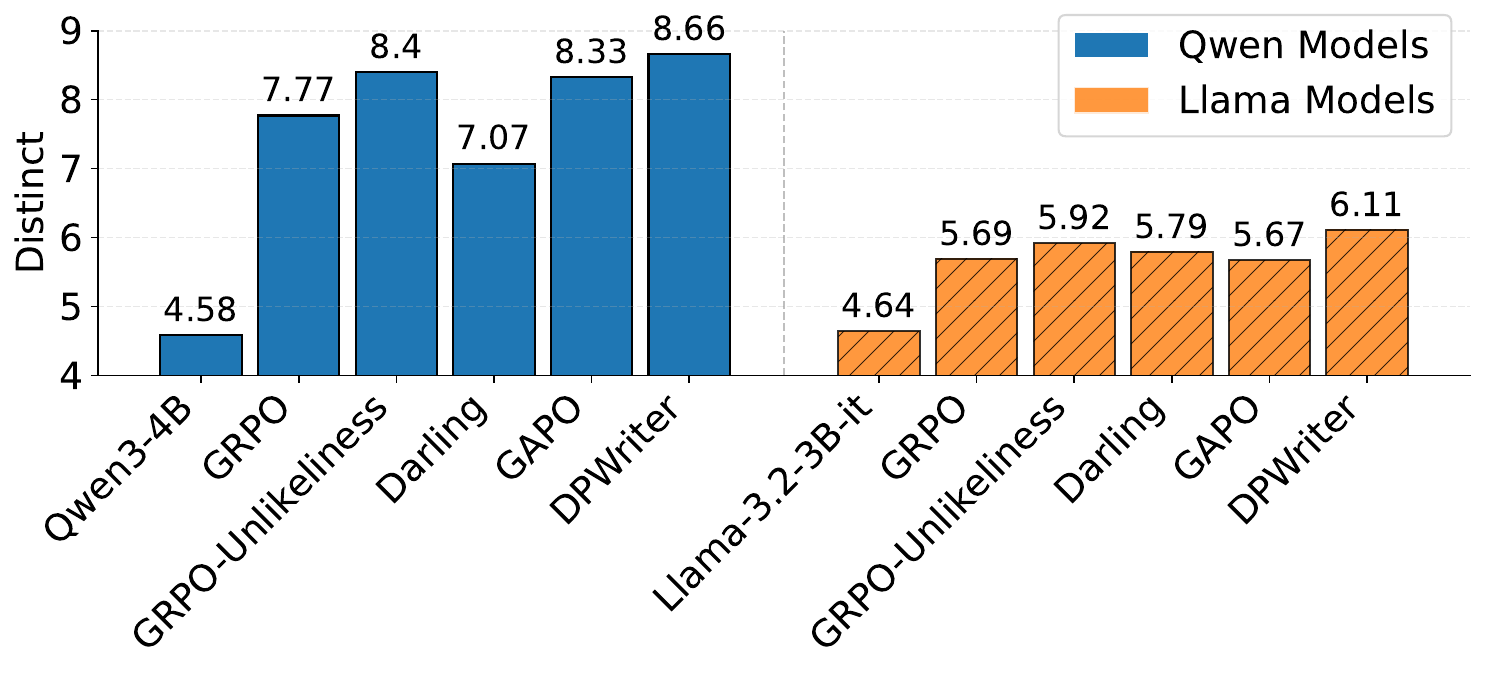}
    \caption{NoveltyBench results comparing \modelname\ with baselines diversity metric Distinct.}
    \label{fig:res_nb_bar}
\end{figure}

\subsection{Main Results}
The main results on the three benchmarks are presented in Table~\ref{tab:main_results}.
Our proposed \modelname\ consistently outperforms all baselines across different backbones on WritingBench and Creative Writing v3 in terms of both quality and diversity metrics.
Notably, on WritingBench, \modelname\ achieves a significant improvement of 15\% in the embedding-based diversity metric and 9.9\% in the EAD metric compared to the standard GRPO method when using the Qwen3-4B backbone.
Meanwhile, this does not come at the cost of quality, as \modelname\ also attains the highest overall score of 6.43.
Results on Creative Writing v3 and ArenaHard v2.0 further validate the effectiveness of our method, with \modelname\ achieving the best diversity metrics and substantial gains in quality metrics.
While GRPO-Unlikeliness achieves competitive performance in embedding-based diversity on ArenaHard v2.0 with a Llama backbone, it lags significantly behind in win rate and EAD metrics.
This suggests that it may exploit embedding-based diversity metrics by generating lower-quality content.

We also evaluate the models on NoveltyBench for diversity assessment, as shown in Figure~\ref{fig:res_nb_bar}.
\modelname\ outperforms all baselines in terms of the Distinct metric, demonstrating its superior capability in generating diverse content.
Consistent results across multiple benchmarks underscore the effectiveness of our proposed \modelname\ in enhancing both the quality and diversity of creative writing.

\begin{table}[t!]
\centering
\scalebox{0.86}{
\setlength{\tabcolsep}{0.4mm}{
\begin{tabular}{lcccccccc}
\toprule
\multicolumn{1}{c}{\multirow{2}{*}{Methods}} &
  \multirow{2}{*}{P} &
  \multirow{2}{*}{T} &
  \multicolumn{3}{c}{\textbf{Qwen3}} &
  \multicolumn{3}{c}{\textbf{Llama3.2}} \\ \cmidrule(lr){4-6} \cmidrule(lr){7-9}
\multicolumn{1}{c}{} &
   &
   &
  \multicolumn{1}{c}{Score} &
  \multicolumn{1}{c}{Emb} &
  \multicolumn{1}{c}{EAD} &
  \multicolumn{1}{c}{Score} &
  \multicolumn{1}{c}{Emb} &
  \multicolumn{1}{c}{EAD} \\ \midrule
DeepWriter      &   & \ding{51} & 6.00 & 10.98 & 8.84 & 4.77 & 12.85 & 8.89 \\
SFT (ours) & \ding{51} & \ding{51} & \textbf{6.04} & 10.99 & \textbf{8.88} & \textbf{4.95} & 12.76 & \textbf{9.10} \\ \midrule
SFT (standard)   &   &   & 5.87 & 10.87 & 8.19 & 4.56 & 12.75 & 8.04 \\ 
\quad w/ think   &   & \ding{51} & 5.93 & \textbf{11.04} & 8.58 & 4.75 & \textbf{12.89} & 8.67 \\
\quad w/ plan    & \ding{51} &   & 5.97 & 10.83 & 8.67 & 4.84 & 12.58 & 8.62 \\
\bottomrule
\end{tabular}}}
\caption{Ablation study on the effects of planning and thinking steps on WritingBench in the SFT stage. `P' and `T' denote planning and thinking steps, respectively.}
\label{tab:ablation_sft}
\end{table}

\begin{table}[t!]
\centering
\scalebox{0.95}{
\setlength{\tabcolsep}{0.7mm}{
\begin{tabular}{lcccc}
\toprule
\multicolumn{1}{c}{\multirow{2}{*}{Methods}} & \multicolumn{3}{c}{\textbf{WritingBench}}      & \textbf{NB} \\ \cmidrule(lr){2-4} \cmidrule(lr){5-5}
\multicolumn{1}{c}{} & Score & Emb   & EAD  & Distinct \\ \midrule
DPWriter                                     & \textbf{6.43} & \textbf{10.45} & \textbf{8.81} & \textbf{8.66}         \\ \midrule
DPWriter-emb         & 6.39  & 10.24 & 8.74 & 8.38     \\
\quad w/o branching        & 6.41  & 10.05 & 8.63 & 8.59     \\
\quad w/o diversity reward & 6.30  & 9.19  & 8.08 & 7.99     \\
GRPO                 & 6.32  & 9.07  & 8.02 & 7.77     \\
\bottomrule
\end{tabular}}}
\caption{Ablation study on the effects of different components in the RL. `NB' denotes NoveltyBench.}
\label{tab:ablation_rl}
\end{table}

\begin{figure*}[t!]
    \centering
    \includegraphics[width=1.0\linewidth]{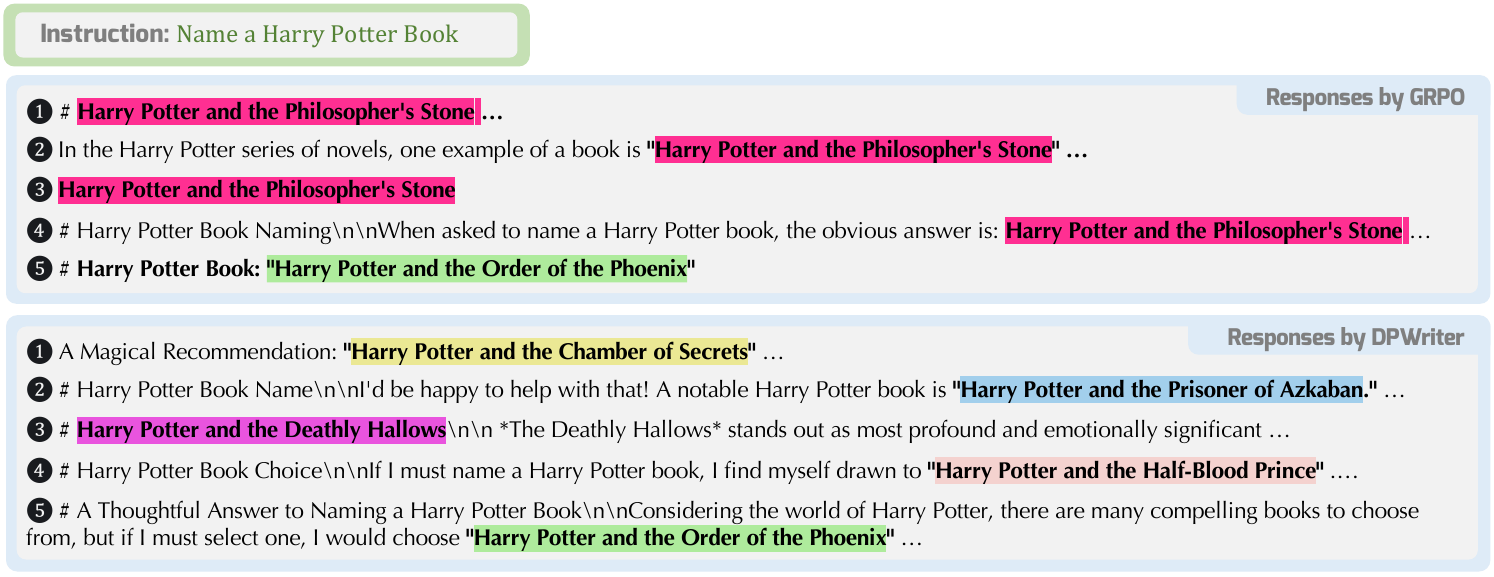}
    \caption{Case study comparing \modelname\ with GRPO on a sample from NoveltyBench. The same answer is highlighted in the same colored background. We present five generations from each model for comparison.}
    \label{fig:case_study}
\end{figure*}

\begin{figure}[t!]
    \centering
    \includegraphics[width=1.0\linewidth]{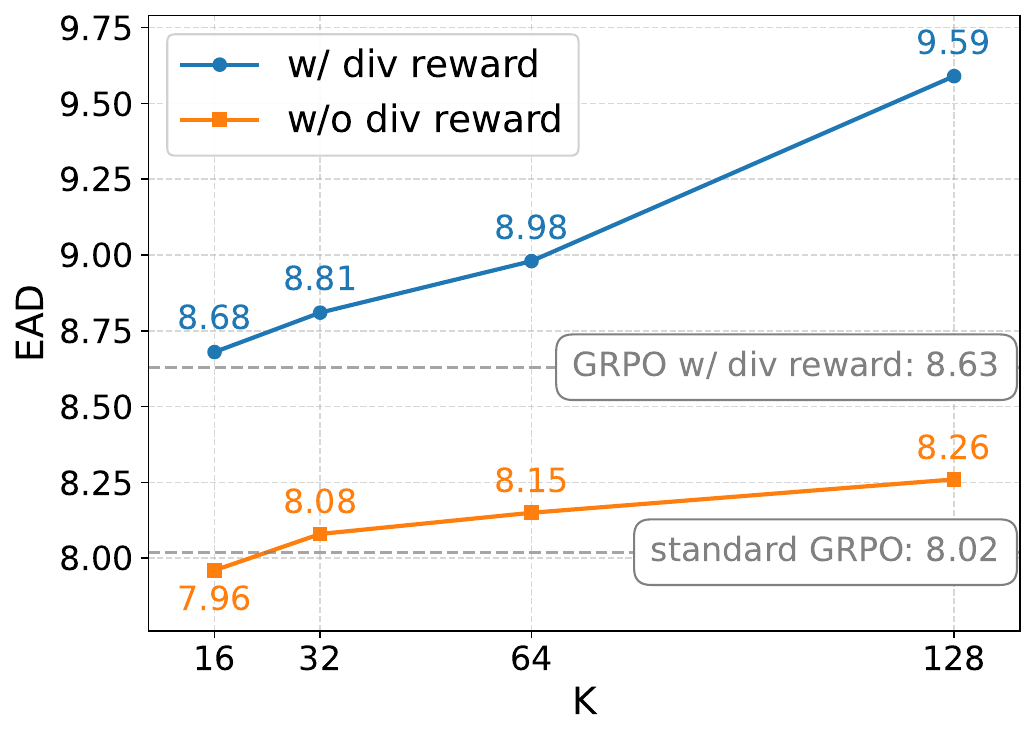}
    \caption{Branching and reward synergy analysis on WritingBench with Qwen3-4B backbone. The "w/ div reward" denotes using diversity reward.}
    \label{fig:branching_reward_synergy}
\end{figure}

\subsection{Ablation Studies}

\paragraph{Ablation on different SFT strategies.}
\label{sec:sft_ablation}
We first investigate the effectiveness of our proposed semi-structured long CoT with planning during cold-start SFT.
We compare our full SFT approach with several variants, including (1) DeepWriter~\cite{wang2025reverse} that synthesizes reasoning trajectories by working ``backwards'' from good responses to discover reasoning processes; (2) standard SFT with long CoT only; (3) SFT with planning only.
As shown in  Table~\ref{tab:ablation_sft}, we observe that with both planning and thinking steps, our SFT model achieves comparable or even better performance than other variants across all metrics and backbones.
This implies that models can better follow the planning-and-thinking paradigm to generate high-quality and diverse content, providing a solid foundation for subsequent RL training.
In addition, our semi-structured long CoT offers more controllability compared to no CoT or standard reasoning trajectories, allowing a diverse branching process for planning generation.

\noindent\textbf{Ablation on different components.}\quad
We further analyze the contributions of different components of our method.
The ablation results are summarized in Table~\ref{tab:ablation_rl}.
We first replace our n-gram-based branching strategy with an embedding-based one (DPWriter-emb).
The performance drops slightly but remains better than the GRPO baseline, indicating the effectiveness of our overall framework.
Next, we remove the branching strategy (w/o branching), which leads to a further decrease in diversity metrics on WritingBench.
This demonstrates that our proposed branching strategy effectively encourages the model to explore diverse content during generation.
Finally, we eliminate the diversity reward (w/o diversity reward), resulting in a significant decline in all diversity metrics, which highlights the importance of explicitly rewarding diversity to guide the model learning.

\noindent\textbf{Branching and Reward in Synergy.}\quad
To further understand the interplay between our branching strategy and diversity reward, we conduct additional experiments to analyze their synergy.
We analyze how diversity metrics vary with different branching factor $K$ in $\{16, 32, 64, 128\}$, with and without the diversity reward.
The results are illustrated in Figure~\ref{fig:branching_reward_synergy}.
We observe that combining branching with the diversity reward consistently yields the highest diversity scores across all values of $K$.
Moreover, as $K$ increases, the diversity metrics also improve, indicating that a larger branching factor allows the model to explore a wider range of content.
When the diversity reward is applied, the diversity curves exhibit a steeper with respect to $K$ than in the absence of the reward. 
This suggests that the branching strategy is more effective in enhancing diversity under the guidance of the diversity reward. 

\subsection{Case Study}
We present a case study in Figure~\ref{fig:case_study} to qualitatively compare the generations from \modelname\ and the GRPO baseline on a sample from NoveltyBench.
From the case study, we observe that \modelname\ generates five distinct responses that cover a wide range of Harry Potter books, while GRPO produces several similar answers, with four out of five generations being ``Harry Potter and the Philosopher's Stone.''
This demonstrates the superior diversity of \modelname\ facilitated by our proposed branching strategy and diversity reward. 
More comparisons are presented in Appendix~\ref{appendix:more_case_study}.

\section{Conclusion}

In this work, we address a fundamental limitation of RL-based enhancement, where improvements in alignment and performance come at the cost of persistent diversity collapse, by introducing a novel semi-structured long chain-of-thought (CoT) reasoning framework.
Our approach explicitly guides diversity exploration through a strategic planning-phase branching mechanism and a group-aware diversity contribution reward design.
Experimental results across multiple creative writing tasks demonstrate that our framework effectively promotes output diversity without sacrificing quality.
The combination of diverse planning and targeted reward signals provides a principled pathway toward more expressive and versatile language generation in open-ended applications.

\section*{Limitations}
While our proposed DPWriter framework effectively enhances output diversity in creative writing tasks, there are several limitations and areas for future improvement.
First, the reliance on a semi-structured CoT and the Diverse Planning Branching method may introduce additional computational overhead, potentially limiting scalability for extremely large models or datasets.
Second, although our work improves diversity without compromising quality, the seesaw between these two aspects may not be fully resolved, and further research is needed to explore more that can better balance them.
Finally, besides quality improvement, whether diversity can benefit other aspects like creativity remains a more open question.

\bibliography{custom}

\appendix

\clearpage

\section{Implementation Details}
\label{appendix:implementation_details}
In RL, the model is trained for 5 epochs, while the max prompt length is set to 1024 and the max response length is set to 3072.
We set the branching factor $K=32$ and use n-gram n-gram-based strategy to evaluate diversity in diverse planning branching.
For reward functions, the diversity weight $\lambda$ is set to 0.6, and quality threshold is $\tau=10$.
For embedding-based similarity calculation, we use the Qwen3-Embedding-0.6B~\cite{zhang2025qwen3} model to extract the embeddings of the generated plans or responses.
During inference, we set the max generation length to 4096 and use nucleus sampling with $p=0.8$ and temperature $T=0.7$, which is aligned with the setting in Writingbench.

\section{Additional Related Works}
\label{appendix:add_related_works}
Investigations into enhancing LLMs' diversity to mitigate mode collapse issues can be broadly categorized into two main lines: inference-time methods and training-time methods.
Besides the training-time methods mentioned in the main text, inference-time diversity methods are also relevant works that we elaborate on below.

\paragraph{Inference-time Diversity.}
To enhance diversity during inference, various methods modify the next-token selection process, enabling quick and flexible generation without altering model parameters. 
A widely adopted approach is to increase the decoding temperature~\cite{Zhu2024Hot,Max2024IsTemp} or to sample from token distributions after applying different cut-off strategies~\cite{Ashwin2016DiverseBeam,Ari2020topp,Minh2024minp,Giorgio2025DiffSampling}.
Some studies focus on prompt engineering to encourage LLMs to draw from broader intent strategies~\cite{Eltayeb2025Intent,Ruan2025G2} or introduce greater randomness into the generation process~\cite{Kou2025String}.
It has been observed that smaller-scale models often exhibit greater output diversity~\cite {Vishakh2024Does}.
Thus, some studies explore collaborative decoding between large and small models, leveraging the diversity of smaller models to enhance the outputs of larger ones~\cite{Lisa2023ContrastiveDecoding,Wang2025Optimizing}.

\begin{table*}[h!]
\centering
\begin{tabular}{p{15cm}}
\toprule
\textbf{Prompt for Generating multi-aspect Plans} \\
\midrule

\#\#\# Task Description \\

You are a writer with meticulous logic and divergent thinking. I will give you a pair of instruction and response, please analyze and summarize the core creativity idea of the response based on the given content, from the following five aspects: \\
1. Goal and audience \\
2. Information and perspective \\
3. Structure and logic \\
4. Language and style \\
5. Presentation and experience \\ \\

Please note: \\
1. Your reply must be complete and contain the above five aspects, and be in JSON format, do not output other content; \\
2. Your reply must reflect the design of the response based on the instruction (as if the response has not been written yet) and avoid summaries or comments on the response; do not include referential phrases such as ``this article'', ``this'', or ``that'', etc.; \\
3. The structure, writing techniques, etc., should be sufficiently specific and detailed. \\
The language of your reply must be consistent with the language of the instruction and response. \\
Format example: \\
\{``xxx'': ``xxx'', ...\} \\ \\

\#\#\# Instruction: \{\{Instruction\}\} \\
\#\#\# Response: \{\{Response\}\} \\
\#\#\# Your reply: \\

\bottomrule
\end{tabular}
\caption{Prompt for Generating multi-aspect Plans.}
\label{app:prompt_generate_plan}
\end{table*}

\begin{table*}[h!]
\centering
\begin{tabular}{p{15cm}}
\toprule
\textbf{Prompt for Generating Plan-consistent CoT} \\
\midrule

\#\#\# Task Description \\
You are a creative writer with logical rigor and divergent thinking. I will provide you with a pair of Instruction and Response, along with a five-dimensional overview of the Response called Plans. Please strictly refer to the content of the Response and Plans, maintain the open format of Think, and modify the content in Think to reflect the five dimensions of Plans, while expanding your thoughts as required. \\ \\

Please note: \\
1. Ensure your response fully includes a thought process in the style of Think, starting directly with mental activity and without outputting any other content. \\
2. Ensure your response is based on the expected thought process for the Instruction (as if the Response has not yet been written), avoiding summaries or comments on the given Response or Think. Do not include direct referential phrases such as ``this piece'', ``the'', or ``this.'' Maintain a tone of experimentation and reflection, avoiding overly technical language. \\
3. Do not directly include the names of each dimension. Instead, reflect the information of each dimension in separate open-form paragraphs. \\
4. Ensure smooth transitions between paragraphs, guiding the thought process naturally and step by step toward the Response. Appropriately diverge and expand on each paragraph. \\
5. During the analysis of each paragraph, identify the key points of the problem and explore multiple angles around these cores. Propose multiple possible response directions or methods. After weighing creative ideas, select one, and ensure the chosen creativity aligns with the Plans. \\
6. If necessary, use ``user'' for address or directly describe the problem. Avoid terms like ``reader.'' \\
7. Ensure the provided response is consistent with the language of the Instruction and Response. \\ \\

\#\#\# Instruction: \{\{Instruction\}\} \\
\#\#\# Response: \{\{Response\}\} \\
\#\#\# Reference Think: \{\{Reference Think\}\} \\
\#\#\# Plans: \{\{Plans\}\} \\
\#\#\# Your reply: \\

\bottomrule
\end{tabular}
\caption{Prompt for Generating Plan-consistent CoT.}
\label{app:prompt_generate_cot}
\end{table*}

\begin{table*}[h!]
\centering
\begin{tabular}{p{15cm}}
\toprule
\textbf{Prompt for Generating Long CoT} \\
\midrule

\#\#\# Task Description \\
You are a logical expert with divergent thinking and rich imagination. I will give you an Instruction and a Response, please refer to their information and give a complete Thought process of Response. \\ \\
Please note: \\
1. Think about the instruction in the form of first-person self-talk, presenting a natural and real psychological process; avoid mechanical listing of points, and use several paragraphs to express; \\
2. Thought should have a clear structure and coherence, strict logic, and progressive, and use reasonable conjunctions to make each layer of Thought naturally connected, and gradually lead to the Response; \\
3. Identify the key points of the problem during the analysis process, brainstorm around these core points from multiple aspects, and propose multiple possible response directions or methods; After giving the optional solutions, the creative ideas need to be verified, and ensure that the ideas selected in Thought are consistent with the final Response; \\
4. Thought should reflect the real creative process insightfully and coherently, showing the uncertainty and trade-offs in exploration, and aim to articulate concise overarching ideas rather than excessive details; \\
5. Avoid directly showing all the information in the Response in Thought, and let each element gradually emerge through natural guidance; \\
6. Make sure the Thought reflects the expected generation (as if the Response has not been written yet), avoid summarizing or commenting on the given Response; do not include phrases such as ``thought process'', ``according to the above prompts'', ``that response'', etc. \\
7. The Thought given must be consistent with the language of the Instruction and Response. \\ \\

\#\#\# Given Instruction: \{\{Instruction\}\} \\
\#\#\# Given Response: \{\{Response\}\} \\
\#\#\# Thought:  \\

\bottomrule
\end{tabular}
\caption{Prompt for Generating Long CoT.}
\label{app:prompt_generate_long_cot}
\end{table*}

\section{Prompts}
\label{appendix:prompts}

We present the prompts we used for constructing multi-aspect plans and generating plan-consistent CoT data in this section.
Table~\ref{app:prompt_generate_plan} is the prompt we use to generate multi-aspect plans and Table~\ref{app:prompt_generate_cot} is the prompt we use to generate plan-consistent CoT, as described in Section~\ref{sec:data_const}.
For those data that are only in the form of instruction-response pairs $(q, y)$ without existing CoT, we use the prompt in Table~\ref{app:prompt_generate_long_cot} to generate long CoT for them.

\section{More Case Studies}
\label{appendix:more_case_study}
We provide more case studies in this section to showcase the effectiveness of our method.
An example from WritingBench is shown in Figure~\ref{fig:case_study.writingbench}.
We can see that \modelname is able to generate high-quality responses by following the multi-aspect plans and plan-consistent CoT.
Another example from NoveltyBench is shown in Figure~\ref{fig:case_study.noveltybench}, which demonstrates that \modelname can produce creative and coherent stories by adhering to the generated plans and thought processes.

\begin{figure*}[htb]
    \begin{tcolorbox}[title=\textbf{\textcolor{white}{An example from WritingBench}}, colframe=darkblue!40!white]

    \textbf{Instruction:} \\
    I need to write a research paper on an AI-based medical imaging-assisted diagnosis system. Please help me draft a paper outline.

    \tcblower

    \textbf{Response of \modelname:} \\
    \texttt{<think>} \\
    To respond creatively, I first conduct a divergent analysis from multiple perspectives and then think carefully step by step. \\
    \texttt{<goal>}Goal and audience: Design an academic research paper outline targeting medical professionals, researchers, and clinicians interested in the application of AI in medical imaging. The goal is to provide a comprehensive framework for understanding, implementing, and evaluating AI-based diagnostic systems, with emphasis on clinical relevance, technical details, and future directions.\texttt{</goal>} \\
    \texttt{<info>}Information and perspective: Integrate multidisciplinary information including medical imaging fundamentals, AI algorithm types, system architecture, data management, clinical workflow integration, evaluation metrics, regulatory considerations, and future research opportunities. Present a balanced perspective that combines technical depth, clinical impact, ethical concerns, and practical implementation challenges. Include both current state-of-the-art examples and future trends to contextualize the field.\texttt{</info>} \\
    \texttt{<struct>}Structure and logic: Organize content into clear, hierarchical sections: Abstract, Introduction (background, problem statement, objectives), Literature Review (traditional methods, AI advancements, comparative analysis), Methodology (system architecture, data acquisition/processing, algorithm selection/training, integration, validation), Results and Discussion (performance metrics, clinical impact, limitations), Future Directions (enhanced algorithms, multimodal fusion, explainability, personalized medicine, deployment strategies), Conclusion (summary, implications, recommendations), Acknowledgments, References. Ensure logical progression from context and background to specific technical details, experimental results, and broader implications. Each section builds upon the previous, supporting a coherent narrative.\texttt{</struct>} \\
    \texttt{<lang>}Language and style: Use formal, precise academic language suitable for peer-reviewed publication. Maintain objectivity and clarity, avoiding colloquialisms. Employ discipline-specific terminology (e.g., convolutional neural networks, transfer learning, precision/recall) and define key concepts where necessary. The style should facilitate comprehension for readers with varying levels of technical expertise, while ensuring rigor and professionalism.\texttt{</lang>} \\
    \texttt{<pres>}Presentation and experience: Create an outline that guides the reader through a structured exploration of the topic, from foundational concepts to advanced applications and future prospects. Facilitate engagement by providing actionable frameworks, real-world case studies, and critical evaluation of both strengths and limitations. Enhance the reading experience with well-labeled sections, bullet points for clarity, and references to standard practices and guidelines. Ensure the presentation supports both quick scanning for key points and deep dives into technical or clinical aspects.\texttt{</pres>} \\

    ... (continued on the next page)
    \end{tcolorbox}
    \caption{An example from WritingBench that are generated by \modelname.}
    \label{fig:case_study.writingbench}
\end{figure*}
\clearpage 
\begin{figure*}[htb]
    \ContinuedFloat 
    \begin{tcolorbox}[title=\textbf{\textcolor{white}{An example from WritingBench}}, colframe=darkblue!40!white]

    ... (continued) \\

    Thinking about how to approach user’s request for a research paper outline on AI-based medical imaging-assisted diagnosis systems, it’s important to clarify the intended audience and purpose first. Medical professionals, researchers, and clinicians are likely seeking both technical rigor and practical guidance, so the outline should be designed to serve those who need to understand not just the theoretical underpinnings but also the real-world applicability and limitations of such systems. The goal becomes one of bridging the gap between cutting-edge AI research and its translation into clinical practice, making sure the framework is comprehensive enough to inform decision-making, yet accessible to readers with diverse backgrounds. \\
    
    To achieve this, the information presented must span multiple domains—medical imaging fundamentals, AI algorithmic approaches, system design, data management, workflow integration, performance evaluation, regulatory compliance, and future research trajectories. It would be valuable to weave together perspectives from computer science, radiology, ethics, and healthcare policy. Including both current best practices and emerging trends allows the outline to reflect the dynamic nature of the field. A balanced viewpoint is crucial, weighing technical advances against issues like bias, interpretability, and privacy, while also highlighting the tangible impacts these systems have on patient outcomes and healthcare economics. \\
    
    Structuring the outline demands careful consideration of logical flow. Starting with an abstract sets expectations and highlights the main contributions. The introduction should establish the context, articulate the clinical and technical problems, and clearly state the objectives. A thorough literature review is essential to situate the work within existing knowledge, contrasting traditional diagnostic methods with contemporary AI solutions and analyzing comparative advantages and drawbacks. Methodology needs to break down the system architecture, detail data acquisition and processing, describe algorithm selection and training, explain integration into clinical workflows, and present validation protocols. Results and discussion should focus on performance metrics, clinical impact, and honest assessment of limitations. Future directions can explore avenues for improved algorithms, multimodal imaging, explainability, personalization, and broader deployment. Finally, the conclusion should synthesize findings, underscore implications, and offer actionable recommendations. Each section must build logically on the last, maintaining coherence and driving the reader toward a holistic understanding. \\
    
    Language choice is another key consideration. Academic formality and precision are paramount, employing discipline-specific terminology without sacrificing clarity. Definitions should be provided for complex concepts, and jargon used judiciously to ensure accessibility for non-specialists while retaining rigor for experts. Objectivity is crucial; claims should be supported by evidence, and limitations acknowledged transparently. The tone should invite critical thinking, encouraging the audience to consider both the promise and the pitfalls of AI in medical imaging. \\

    ... (continued on the next page)
    \end{tcolorbox}
    \caption{An example from WritingBench that are generated by \modelname\ (continued).}
    \label{fig:case_study.writingbench.continued1}
\end{figure*}
\clearpage 
\begin{figure*}[htb]
    \ContinuedFloat 
    \begin{tcolorbox}[title=\textbf{\textcolor{white}{An example from WritingBench}}, colframe=darkblue!40!white]
    ... (continued) \\
    
    The presentation itself should be designed for user engagement and navigation. Well-labeled sections, bullet points for clarity, and references to standard practices enhance readability. Including real-world case studies or examples can ground the discussion in practical realities, while action-oriented frameworks guide implementation. Critical evaluation—highlighting both strengths and weaknesses—is vital to foster trust and informed decision-making. By guiding the reader from foundational concepts to advanced topics and future possibilities, the outline becomes not just a summary but a roadmap for both research and clinical adoption, facilitating both rapid overview and deeper exploration as needed. \\
    \texttt{</think>} \\

    \# Research Paper Outline: AI-Based Medical Imaging-Assisted Diagnosis System \\

    \#\# 1. Abstract \\
    - Brief summary of the research focus, methodology, key findings, and significance \\
    - Highlight the potential impact of AI in medical imaging diagnostics \\
    - Mention any novel contributions or innovations in the proposed system \\

    \#\# 2. Introduction \\
    \#\#\# 2.1 Background and Significance \\
    - Current challenges in medical imaging diagnosis \\
    - Limitations of traditional diagnostic methods \\
    - Growing role of artificial intelligence in healthcare \\
    - Statistics on diagnostic errors and their consequences \\

    \#\#\# 2.2 Research Objectives \\
    - Primary goal of developing an AI-assisted diagnosis system \\
    - Specific objectives to be addressed \\
    - Expected benefits for clinicians and patients \\

    \#\#\# 2.3 Scope and Limitations \\
    - Definition of the problem space \\
    - Constraints of the current research \\
    - Applicability of findings to other medical imaging modalities \\

    \#\# 3. Literature Review \\
    \#\#\# 3.1 Traditional Medical Imaging Techniques \\
    - Overview of common imaging modalities (X-ray, CT, MRI, ultrasound, etc.) \\
    - Evolution of image interpretation techniques \\
    - Human factors in medical image analysis \\

    ... (continued on the next page)
    \end{tcolorbox}
    \caption{An example from WritingBench that are generated by \modelname\ (continued).}
    \label{fig:case_study.writingbench.continued2}
\end{figure*}
\clearpage 
\begin{figure*}[htb]
    \ContinuedFloat 
    \begin{tcolorbox}[title=\textbf{\textcolor{white}{An example from WritingBench}}, colframe=darkblue!40!white]
    ... (continued) \\
    
    \#\#\# 3.2 Artificial Intelligence in Medical Imaging \\
    - Types of AI algorithms applied in medical imaging \\
    - Convolutional Neural Networks \\
    - Deep Learning architectures \\
    - Computer Vision techniques \\
    - Major breakthroughs and recent advancements \\
    - Comparative analysis of different AI approaches \\

    \#\#\# 3.3 Existing AI-Assisted Diagnosis Systems \\
    - Comprehensive review of current commercial and research-based systems \\
    - Performance metrics and clinical outcomes \\
    - Integration challenges with existing healthcare workflows \\

    \#\#\# 3.4 Knowledge Gaps and Opportunities \\
    - Areas requiring further research \\
    - Current limitations in AI-based medical imaging \\
    - Emerging trends and future directions \\

    \#\# 4. Methodology \\
    \#\#\# 4.1 System Architecture \\
    - Overall design of the AI-assisted diagnosis platform \\
    - Modular components and their interactions \\
    - Data flow and processing pipeline \\

    - User interface design for clinicians \\

    \#\#\# 4.2 Data Acquisition and Preprocessing \\
    - Dataset selection criteria and sources \\
    - Image acquisition protocols \\
    - Preprocessing techniques (normalization, enhancement, segmentation) \\
    - Data augmentation strategies \\
    - Privacy-preserving data handling methods \\

    \#\#\# 4.3 Algorithm Development and Selection \\
    - Choice of AI model architecture \\
    - Training methodology and optimization techniques \\
    - Hyperparameter tuning approach \\
    - Transfer learning applications \\
    - Model validation and cross-validation strategies \\

    ... (continued on the next page)
    \end{tcolorbox}
    \caption{An example from WritingBench that are generated by \modelname\ (continued).}
    \label{fig:data_example.lyrics.continued3}
\end{figure*}
\clearpage 
\begin{figure*}[htb]
    \ContinuedFloat 
    \begin{tcolorbox}[title=\textbf{\textcolor{white}{An example from WritingBench}}, colframe=darkblue!40!white]
    ... (continued) \\
    
    \#\#\# 4.4 Integration with Clinical Workflows \\
    - Interface design with existing PACS (Picture Archiving and Communication Systems) \\
    - Workflow optimization for radiologists \\
    - Alert generation and prioritization mechanisms \\
    - Decision support level (detection, diagnosis, treatment planning) \\

    \#\#\# 4.5 Evaluation Framework
    - Performance metrics (accuracy, sensitivity, specificity, AUC, etc.) \\
    - Comparison with expert human performance \\
    - Clinical utility assessment \\
    - Validation on independent datasets \\
    - Ethical considerations in evaluation \\

    \#\# 5. Results and Discussion \\
    \#\#\# 5.1 System Performance Analysis \\
    - Quantitative results of AI model performance \\
    - Comparative analysis with traditional methods \\
    - Statistical significance of improvements \\
    - Error analysis and case studies \\

    \#\#\# 5.2 Clinical Impact Assessment \\
    - Effectiveness in improving diagnostic accuracy \\
    - Time efficiency gains for clinicians \\
    - Reduction in false positives/negatives \\
    - Patient outcome improvements \\

    \#\#\# 5.3 Implementation Challenges \\
    - Technical integration hurdles \\
    - Clinician acceptance and training requirements \\
    - Regulatory approval process \\
    - Data security and privacy concerns \\

    \#\#\# 5.4 Limitations and Future Work \\
    - Identified weaknesses in the current system \\
    - Potential biases and generalizability issues \\
    - Required infrastructure upgrades \\
    - Next steps for development and refinement \\

    \#\# 6. Future Directions \\
    \#\#\# 6.1 Advanced Algorithmic Approaches \\
    - Exploration of more sophisticated neural network architectures \\
    - Reinforcement learning applications \\
    - Federated learning for collaborative model improvement \\
    - Explainable AI techniques \\
    
    ... (continued on the next page)
    \end{tcolorbox}
    \caption{An example from WritingBench that are generated by \modelname\ (continued).}
    \label{fig:data_example.lyrics.continued4}
\end{figure*}
\clearpage 
\begin{figure*}[htb]
    \ContinuedFloat 
    \begin{tcolorbox}[title=\textbf{\textcolor{white}{An example from WritingBench}}, colframe=darkblue!40!white]
    ... (continued) \\
    
    \#\#\# 6.2 Multimodal Imaging Integration \\
    - Combining different imaging modalities for complementary insights \\
    - Real-time fusion of data streams \\
    - Cross-modality transfer learning \\
    - Personalized imaging protocols \\

    \#\#\# 6.3 Clinical Deployment Strategies \\
    - Scalable implementation across different healthcare settings \\
    - Mobile and edge computing applications \\
    - Continuous learning and adaptation mechanisms \\
    - Cost-effectiveness analysis \\

    \#\#\# 6.4 Ethical and Regulatory Considerations \\
    - Addressing algorithmic bias and fairness \\
    - Maintaining physician autonomy \\
    - Evolving regulatory frameworks \\
    - Global implementation challenges \\

    \#\# 7. Conclusion \\
    \#\#\# 7.1 Summary of Findings \\
    - Recapitulation of key research outcomes \\
    - Contribution to the field of medical AI \\
    - Validation of initial hypotheses \\

    \#\#\# 7.2 Implications for Practice \\
    - Recommendations for clinical implementation \\
    - Training requirements for healthcare professionals \\
    - Infrastructure investments needed \\
    - Policy changes potentially required \\

    \#\#\# 7.3 Final Thoughts \\
    - Reflection on the transformative potential of AI in diagnostics \\
    - Balancing automation with human judgment \\
    - Ethical responsibilities in AI deployment \\
    - Call to action for continued research and collaboration \\

    \#\# 8. Acknowledgments \\
    - Recognition of contributors, funding sources, and institutions \\

    ... (continued on the next page)
    \end{tcolorbox}
    \caption{An example from WritingBench that are generated by \modelname\ (continued).}
    \label{fig:data_example.lyrics.continued4}
\end{figure*}
\clearpage 
\begin{figure*}[htb]
    \ContinuedFloat 
    \begin{tcolorbox}[title=\textbf{\textcolor{white}{An example from WritingBench}}, colframe=darkblue!40!white]
    ... (continued) \\

    \#\# 9. References \\
    - Comprehensive list of academic sources, industry reports, and relevant publications \\
    - Follow standard citation format (e.g., APA, IEEE, Chicago) \\

    --- \\

    This outline provides a comprehensive structure for a research paper on AI-based medical imaging-assisted diagnosis systems. It balances technical depth with clinical relevance, addresses both current capabilities and future directions, and incorporates ethical and practical considerations essential for successful implementation in healthcare settings.

    \end{tcolorbox}
    \caption{An example from WritingBench that are generated by \modelname\ (continued).}
    \label{fig:data_example.lyrics.continued4}
\end{figure*}

\begin{figure*}[htb]
    \begin{tcolorbox}[title=\textbf{\textcolor{white}{An example from NoveltyBench}}, colframe=darkblue!40!white]

    \textbf{Instruction:} \\

    \tcblower

    \textbf{Response of \modelname:} \\

    \texttt{<think>} \\
    To respond creatively, I first conduct a divergent analysis from multiple perspectives and then think carefully step by step. \\
    \texttt{<goal>}Goal and audience: The goal is to craft a concise, emotionally resonant narrative featuring a girl and her dog, targeting a general audience seeking short, heartfelt storytelling. The story is designed to evoke empathy and warmth, appealing to readers who appreciate character-driven, endearing tales.\texttt{</goal>} \\
    \texttt{<info>}Information and perspective: The story centers on a girl and her dog as they navigate a challenging, possibly rainy journey together. The perspective is close, focusing on their shared experience and emotional connection. Information is presented as a sequence of events, with subtle hints at personal growth and mutual support.\texttt{</info>} \\
    \texttt{<struct>}Structure and logic: The response employs a five-sentence structure, each sentence advancing the plot and deepening the emotional arc. The logical flow transitions from the setting and preparation, through the journey and emotional struggle, to a resolution and a personal epiphany. The narrative builds toward a cathartic conclusion, aligning with classic three-act structure within the sentence limit.\texttt{</struct>} \\
    \texttt{<lang>}Language and style: The language is straightforward yet evocative, combining direct description with internal reflection. Dialogue and action are used sparingly to focus on the emotional journey. The style leans toward sentimental realism, using simple vocabulary and present-tense narration for immediacy and intimacy.\texttt{</lang>} \\
    \texttt{<pres>}Presentation and experience: The presentation is minimalist, relying on vivid imagery and emotional subtext to engage the reader. The brevity of the format encourages reflection, with each sentence contributing to a complete story arc. The experience is designed to be relatable and comforting, leaving readers with a sense of hope and companionship.\texttt{</pres>} \\
    Considering the challenge, my mind immediately begins with a focus on what resonates emotionally with the user-who seeks a story in five sentences about a girl and her dog. There's an expectation of warmth and a universal appeal; the piece should feel inviting and familiar, perhaps even cathartic. I imagine who might be drawn to such a story: someone longing for a gentle comfort, or looking for a slice of daily life that's both relatable and rich in feeling. \\

    The relationship at the heart of the story is crucial. Should the narrative be about their adventures, their care, or a moment of quiet intimacy? I could lean toward the dog as companion, trotting alongside the girl, whether through nature or on a stormy day, each step together representing trust and support. Maybe the girl's struggles, and the dog's loyality, highlight a bond that transforms hardship into strength. This raises the possibility of exploring not just the action, but the girl's emotional state-the way she leans on her dog in moments of adversity. \\

    ... (continued on the next page)
    \end{tcolorbox}
    \caption{An example from NoveltyBench that are generated by \modelname.}
    \label{fig:case_study.noveltybench}
\end{figure*}
\clearpage 
\begin{figure*}[htb]
    \ContinuedFloat 
    \begin{tcolorbox}[title=\textbf{\textcolor{white}{An example from NoveltyBench}}, colframe=darkblue!40!white]

    ... (continued) \\
    
    Structurally, each sentence must matter, ever inch of the journey advancing the story. I need to decide whether the tale should flow seamlessly from a realization about the weather, through their preparation and encounter, into the respite of inside, and end with a sense of acceptance or epiphany. Could the story be set during a storm, each sentence tightening the connection between outside peril and inside comfort, and the final image offering closure? I'm drawn to the classic structure-encounter, tension, action, resolution, and realization-confined within five sentences but with enough depth to leave an impact. \\

    The language must be clear and evocative, with enough detail to suggest the scene without bogging down the pace. I might use simple, direct sentences, trading explicit dialogue for gestures and actions-perhaps the girl closing her umbrella, the dog basking in warmth, her eyes softening in relief. Sentimental realism, perhaps with present-tense narration, could root the story in the everyday, heightening its feeling of immediacy and intimacy. \\

    Through this, I aim for presentation that is spare yet vivid, each line a window into the shared world of the girl and her dog. The experience should linger-the warmth of companionship after rain, the gentle shift from anxiety to hope, leaving the user with a sense of comfort and perhaps reflection on their own relationships. Considering these avenues, I settle on a scenario where the story's emotional weight lies in how the girl and her dog find solace together, the journey completed not just in reaching an end, but in finding each other along the way. \\
    \texttt{</think>}

    \# Five Sentences: Girl and Dog \\

    \#\# Version 1: \\

    The rain fell steadily as Emma adjusted her umbrella, knowing Max would insist on staying outside until the sky cleared. When she finally scooped him up, his wet fur glistened against her coat. Entering the shelter, Emma felt the weight lifting from her shoulders. She realized she'd missed how much she loved walking with Max, not just living with him. Shelter complete, so was her heart's journey. \\

    \#\# Version 2 (More direct): \\

    Scanning the weather report, Hannah decided to cancel her afternoon hike. Max whined, persistent despite her decision. Reluctantly, she opened the door and placed him in the hallway. Inside the warmth, Max stretched and curled up beside her, purring softly. Something shifted in Hannah's shoulders; she never needed Max to keep her dry, just to remind her where it was possible to be safe. \\

    ... (continued on the next page)
    \end{tcolorbox}
    \caption{An example from NoveltyBench that are generated by \modelname\ (continued).}
    \label{fig:case_study.noveltybench.continued1}
\end{figure*}
\clearpage 
\begin{figure*}[htb]
    \ContinuedFloat 
    \begin{tcolorbox}[title=\textbf{\textcolor{white}{An example from NoveltyBench}}, colframe=darkblue!40!white]

    ... (continued) \\
    
    \#\# Version 3 (Dark twist): \\

    Lila's umbrella stayed closed, protecting her from the pouring rain. Across the street, Max trotted home, his collar drenched but his spirit untouched. When Lila made it inside, she found Max waiting by the fireplace, his damp fur twitching with anticipation. Despite the storm, despite her panic attack, the simple gesture of a waiting dog allowed Lila to breathe again. This is where I belong, she thought. \\

    Considering different approaches, I was drawn to a story that starts with the girl anticipating something outside (a walk, a hike, or weather-related plan) and having to choose between her commitment or her dog's loyalty. \\

    The middle portion should focus on the emotional tension-her reluctance versus her dog's enthusiasm or need. \\

    The ending needs to capture the transformation: whether catharsis, connection, or recognition of what truly matters. Each version meets the five-sentence constraint while showing depth and emotion, with the final version providing a subtle message about belonging and what's truly important. \\

    \end{tcolorbox}
    \caption{An example from NoveltyBench that are generated by \modelname\ (continued).}
    \label{fig:case_study.noveltybench.continued2}
\end{figure*}

\end{document}